\documentclass{IEEEtran}
\usepackage[T1]{fontenc}
\usepackage{cite}
\usepackage{amsmath,amssymb,amsfonts}
\usepackage{algorithmic}
\usepackage{graphicx}
\usepackage{textcomp}

\usepackage{listings}
\usepackage{xcolor}
\usepackage{geometry}
\usepackage{fancyvrb}
\usepackage{url}
\usepackage{hyperref}
\usepackage{minted}
\UseRawInputEncoding
\usepackage{bm}

\lstdefinestyle{json}{
    basicstyle=\tiny\ttfamily,
    numbers=left,
    numberstyle=\tiny\color{gray},
    stepnumber=1,
    numbersep=5pt,
    showstringspaces=false,
    breaklines=true,
    breakatwhitespace=false,
    frame=single,
    tabsize=2,
    captionpos=b,
}

\def\BibTeX{{\rm B\kern-.05em{\sc i\kern-.025em b}\kern-.08em
T\kern-.1667em\lower.7ex\hbox{E}\kern-.125emX}}

\begin{document}

\title{\textsc{MultiSeismo}: A Multimodal Seismic Dataset and Model for Cross-Modal Seismic Understanding}

\author{Sai~Munikoti,~\IEEEmembership{Member,~IEEE,}
        Ian~Stewart,
        Chengping~Chai,
        Lisa~Linville,
        Scott~Vasquez,
        Sameera~Horawalavithana,
        and~Karl~Pazdernik%
\thanks{This work was supported by the NNSA Office of
Defense Nuclear Nonproliferation Research and
Development, U.S. Department of Energy; in part by
Pacific Northwest National Laboratory, which is operated by Battelle Memorial Institute for the U.S. Department of Energy under contract DEAC05–76RLO1830; in part by the UT-Battelle, LLC, under contract DE-AC05-00OR22725 with the US Department of Energy (DOE). This manuscript is cleared by PNNL for public release: PNNL-SA-221837. This work has been submitted to the IEEE for possible publication. Copyright may be transferred without notice, after which this version may no longer be accessible.}
\thanks{S. Munikoti, I. Stewart, S. Vasquez, S. Horawalavithana, and K. Pazdernik are with Pacific Northwest National Laboratory (PNNL), Richland, WA 99354 USA.}
\thanks{C. Chai is with Oak Ridge National Laboratory (ORNL), Oak Ridge, TN 37830 USA.}
\thanks{L. Linville is with Sandia National Laboratory (SNL), Albuquerque, NM 87123 USA.}
\thanks{K. Pazdernik is also with North Carolina State University (NCSU), Raleigh, NC 27695 USA.}
\thanks{S. Munikoti and I. Stewart contributed equally to this work.}
\thanks{S. Vasquez: Work done when author was at PNNL.}
}

\markboth{Munikoti \MakeLowercase{\textit{et al.}}: MultiSeismo: A Multimodal Seismic Dataset and Model for Cross-Modal Seismic Understanding}
{Munikoti \MakeLowercase{\textit{et al.}}: MultiSeismo: A Multimodal Seismic Dataset and Model for Cross-Modal Seismic Understanding}

\newcommand{\fix}{\marginpar{FIX}}
\newcommand{\new}{\marginpar{NEW}}
\newcommand{\multiseismo}{\textsc{MultiSeismo}}
\newcommand{\micse}{\textsc{MICSE}}
\newcommand{\revone}[1]{\textcolor{black}{#1}}

\maketitle
\begin{abstract}
The application of generalist multimodal models (GMMs) to specialized scientific domains remains limited due to the scarcity of comprehensive domain-specific datasets that integrate multiple data modalities beyond text and images. In seismology, understanding earthquake phenomena requires the synthesis of time-series waveform data, geographical imagery, and contextual metadata, a multimodal integration absent in existing seismic datasets. We present \textsc{MultiSeismo}, a large-scale structured multimodal seismic dataset, comprising over $16K$ seismic events spanning 13 years (2010-2023) across diverse geographical regions. Each event data integrates waveform recordings from global station networks, intensity maps, population exposure visualizations, and a comprehensive textual description within a standardized JSON format. We additionally develop MISCE, a multimodal instruction set on top of raw data to enable supervised training and evaluation of GMMs on seismic reasoning tasks ranging from basic information retrieval to complex cross-modal analysis. 
\revone{
We leverage MISCE to fine-tune an existing multimodal model (Unified-IO 2) enhanced with a specialized time-series encoder, which yields SeisModal, the first domain-specific multimodal model for comprehensive seismic analysis.
Evaluation of state-of-the-art multimodal models on \textsc{MultiSeismo} reveals significant challenges, particularly with time-series data processing for general-purpose models, while demonstrating SeisModal's superior performance on seismic multimodal reasoning tasks. 
These results prove that  \textsc{MultiSeismo} provides a rigorous benchmark for future multimodal research in seismology and validate the success of our domain-specific architectural adaptations. This work offers essential resources and methodological frameworks that can advance multimodal AI applications in seismology and provides transferable approaches for other specialized scientific domains.}
\end{abstract}


\section{Introduction}
\label{sec:introduction}

Research in machine learning shows an increasing interest of generalist multimodal models that can learn and understand across multiple modalities beyond standard text and vision \cite{munikoti2026generalist}, such as video, audio, and time series. 
The rapid advancement of generalist multimodal models (GMMs) has demonstrated remarkable capabilities across diverse domains, from medical to autonomous driving.
However, the application of GMMs to specialized scientific domains remains limited, primarily due to the scarcity of domain-specific multimodal datasets and models that go beyond text and images. 

Among other scientific fields, seismology presents an unique challenge for multimodal AI.
Understanding the origin and impact of seismic events often requires the integration of time series waveform data, spatial geographical information (e.g., satellite images, subsurface models, or topographical maps), and contextual metadata in text form. 
Conventional datasets~\cite{woollam2022seisbench,mousavi2019stanford} and  models~\cite{wang2025seismollm} available for seismic analysis often feature individual modalities, often only sharing time series data without alignment to other modalities.
From a GMM development perspective, the integration of waveform, image, and textual metadata would enable a more holistic view of seismic events both for analysis and model training.


\begin{figure*}[h!]
\centering
\includegraphics[width=0.95\textwidth]{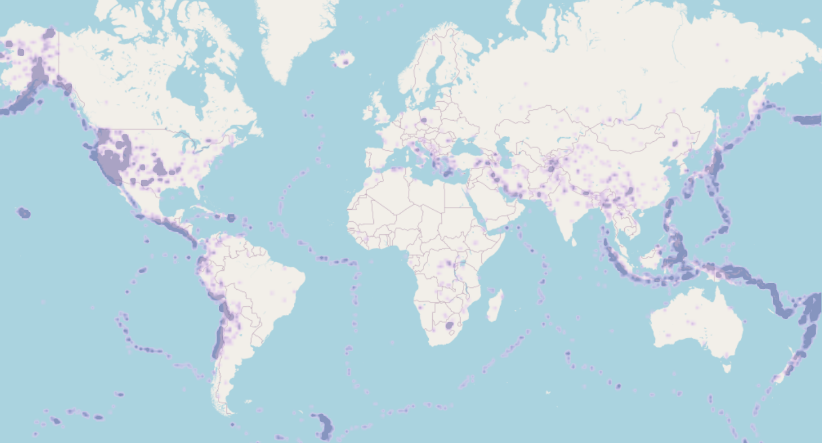}
\caption[short]{Spatial density of the seismic events in the \multiseismo{} dataset (darker = more frequent).}
\label{fig:diagram:spatial_distribution}
\end{figure*}

To address this critical gap, we develop a comprehensive multimodal seismic dataset by systematically curating data from the USGS National Earthquake Information Center (NEIC) \cite{USGS_ShakeMap}: \multiseismo{} (\textbf{Multi}modal \textbf{Seis}mic Dataset for Cross-\textbf{Mo}dal Seismic Understanding). 
Our data collection methodology includes raw waveform recordings from hundreds of stations worldwide; technical and geographical details in text; and images that show intensity, population exposure, and sensor recordings in map form. 
\multiseismo{} includes over 16000 events comprising 13 years of seismic activity (2010-2023) across diverse geographical regions, and it includes events of over 2.5 magnitude and varying depths, which ensures broad representation of global seismic phenomena. Each data sample representing a seismic event integrates all data modalities within a standardized JSON format, transforming traditionally fragmented seismic data modalities into a unified structure that facilitates seamless training and evaluation of GMMs.
To ensure reproducibility and enable systematic dataset expansion, we established a standardized data processing pipeline during \multiseismo{} development that leverages the consistent format of input data from NEIC, facilitating seamless future integration of additional seismic events into the dataset. \multiseismo{} dataset can be found here \url{https://huggingface.co/datasets/PNNL/MultiSeismo}.


Leveraging the standardized data in \multiseismo{}, we also develop a comprehensive set of model-ready instructions, \micse{} (\textbf{M}ultimodal \textbf{I}nstructions to \textbf{C}haracterize \textbf{S}eismic \textbf{E}vents) that enables training and evaluation of GMMs on seismic tasks.
These instructions test models' seismic reasoning ability and encompass various complexity levels, ranging from basic event information retrieval and waveform interpretation to complex reasoning tasks involving image analysis and temporal pattern recognition.

\revone{
Beyond dataset creation, we make a methodological contribution by augmenting an existing multimodal model (i.e., Unified-IO 2) with a specialized time series encoder designed to handle seismic waveform data effectively. We train this enhanced architecture on MICSE, resulting in seismic-domain multimodal model  \textbf{SeisModal}, the first generalist multimodal model tailored for comprehensive seismic analysis. SeisModal demonstrates how general-purpose multimodal architectures can be adapted for specialized scientific domains through targeted architectural modifications and domain-specific training.}


To demonstrate the utility of the dataset, we test the seismic reasoning ability of multiple GMMs, including SeisModal, Unified-IO 2~\cite{lu2024unified}, and Phi-4~\cite{abouelenin2025phi4}. We test cross-modality reasoning (\ref{sec:instruction_evaluation}), handling increasingly long time series sequences (\ref{sec:time_series_length_evaluation}), and retrieving multimodal data (\ref{sec:multimodal_retrieval}). 
\revone{
Our evaluation reveals that SeisModal outperforms general-purpose models on seismic multimodal reasoning tasks, validating both the quality of our dataset and the effectiveness of domain-specific model adaptations.}
Based on low scores across the board, particularly in the time series modality, we determine that the tasks are challenging but tractable, thus providing a strong benchmark on the curated \multiseismo{} dataset to judge multimodal seismic reasoning ability in future models.
\revone{This work thus provides both a comprehensive multimodal dataset and a specialized model for seismology researchers, while offering valuable insights into the development of domain-specific generalist multimodal AI systems.}

\section{Related Work}
\label{sec:related_work}

Seismology is a widely studied topic in the AI community, specifically around the understanding and characterization of time series waveform data. One of the most prominent efforts in this space is SeisBench ~\cite{woollam2022seisbench}, which provides an extensive suite of seismic datasets and tools customized for benchmarking AI algorithms in seismology. SeisBench consolidates multiple well-established seismic datasets, including STEAD (STanford EArthquake Dataset; \cite{mousavi2019stanford}), INSTANCE (Italian Seismic Dataset; \cite{michelini2021instance}), ETHZ, GEOFON, LENDB, NEIC, and SCEDC, among others, each serving a unique analytical purposes ranging from earthquake detection and phase picking to ground motion prediction. The NEIC dataset within SeisBench contains a curated subset of seismic data from the National Earthquake Information Center~\cite{guy2015national}, the same source of the data for this study. In total, NEIC comprises ~1.3 million seismic phase arrivals with global source-station paths as of 2025. However, it doesn't contain the information on the trace start-time and station information. Similarly, the STEAD dataset~\cite{mousavi2019stanford} within SeisBench contains local seismic signals, both earthquake and non-earthquake, along with noise examples. In total it comprises approximately 1.2 million time series, of which around 100K are noise examples and the remaining contain seismic event signals. Overall, SeisBench primarily contains earthquake event metadata and associated parameters optimized for traditional seismological tasks such as arrival time picking and event classification.

Existing benchmark datasets mainly focus on well-defined seismic analysis tasks such as waveform classification, P/S wave arrival time picking, and event detection, which have faciliated reproducible research across the seismological community~\cite{munchmeyer2022picker,zhang2022loc,zhu2023seismic}. 
The diversity of geographical coverage, from regional stations to global networks, and the variety of seismic phenomena represented have established SeisBench in particular as a foundational resource for developing and benchmarking deep learning models in seismology.
However, the existing seismic datasets are predominantly unimodal, focusing primarily on waveform data with accompanying metadata.
Such datasets lack the integrated multimodal structure necessary for training and evaluating cross-modal AI models, particularly lacking text data and image data to contextualize the waveforms (cf. social media earthquake data \cite{mousavi2025gemini}).
Furthermore, the absence of natural language instructions in seismic datasets limits their applicability for training multimodal language models that can interpret, reason about, and communicate findings in human-understandable formats~\cite{abouelenin2025phi4}.

\revone{
Recent advances in multimodal models have demonstrated remarkable capabilities in integrating and reasoning across diverse data modalities, extending beyond traditional text and vision to include audio, video, and specialized scientific data types \cite{munikoti2026generalist,han2023onellm, wu2024next}. Specifically, in the scientific domain, several efforts have begun exploring multimodal AI applications, though most remain limited to text and image modalities. Some notable work includes multimodal scientific question answering \cite{li2024m3sciqa,lu2022learn} , medical imaging with textual reports \cite{ye2025multimodal}, and astronomical data analysis combining images with catalog information \cite{audenaert2024multimodal}. Time-series integration in multimodal models remains particularly challenging, with recent approaches 
such as Time-LLM, demonstrate that by "reprogramming" large LLMs with temporal embeddings, these systems can align numerical sequences with natural language \cite{jin2023time}. Furthermore, vision-centric approaches like Time-VLM treat time-series signals as visual patterns, leveraging the pre-trained spatial reasoning of Vision-Language Models to enhance cross-modal feature extraction and robustness against noisy data \cite{zhong2025time,daswani2024plots}. However, no existing work has comprehensively addressed the unique challenges of seismic data, which requires simultaneous understanding of temporal waveform patterns, spatial geographical context, and domain-specific textual information. This gap underscores the need for specialized datasets and models that can effectively handle the multimodal nature of seismological analysis. Our work fills this critical need for truly multimodal seismic datasets and seismic multimodal model that bridge the gap between traditional seismological data analysis and emerging AI capabilities.}
\section{Seismic Multimodal dataset }
\label{sec:seismultimodal}

This section provides a comprehensive overview of the \multiseismo{} dataset development pipeline. Following subsections detail our systematic approach to data collection from NEIC repository  \cite{USGS_ShakeMap}, describe the preprocessing and aggregation procedures applied to ensure the dataset integrity and standardization, and present the instruction generation framework designed to enable comprehensive evaluation of GMMs on seismic analysis tasks.

\subsection{Raw dataset collection}

We collect raw seismic data from the USGS NEIC shakemap archive. The timeline of the seismic events spans from January 1, 2010 to December 31, 2023, covering nearly 14 years. We capture all events with a magnitude of 2.5 and above from around the world. The majority of these events (99\%) are earthquakes, with only a small number attributed to mining explosions, volcanic eruptions, and other non-tectonic causes. 

\begin{table*}[h]
\caption{Event page information.}
\centering
\begin{tabular}{|c|c|p{8cm}|}
\hline
\textbf{Event page}  & \textbf{Download format} & \textbf{Target content} \\ \hline
Regional Information & HTML                     & Context about the tectonic setting and seismic history of the earthquake's location \\ \hline
PAGER                & PDF                      & Brief estimates of earthquake impact on population and infrastructure                                                                                                                             \\ \hline
Origin               & HTML                     & Fundamental earthquake parameters: location (latitude, longitude, depth), magnitude, and origin time                                                                                                                                    \\ \hline
Moment Tensor        & HTML                     & Earthquake's fault mechanism - how the fault moved during rupture                                                                                                                                  \\ \hline
ShakeMap             & PNG                      & Geographic distribution of ground shaking intensity across the affected region                                                                                                                                    \\ \hline
Waveforms            & HDF5                     & Seismometer recordings from nearby monitoring stations                                                                                                                                   \\ \hline
\end{tabular}
\label{tab:eventpage}
\end{table*}

Except waveforms, all other event data is scraped from USGS ShakeMap Archives \cite{USGS_ShakeMap}. For each seismic event, we scraped multiple HTML pages and downloaded the content as detailed in Table \ref{tab:eventpage}. To download event-specific waveform data, we leveraged the web services provided by National Science Foundation National Geophysical Facility IRIS \cite{IRIS_Wilber3} and the Obspy package.
We only collected data from events that are available in USGS ShakeMap Archives and IRIS.

\subsection{Data post processing and Analysis}
The majority of collected text and image data exists in HTML and PDF formats, containing substantial extraneous information alongside the target content relevant for model training/evaluation and for seismic analysis applications.
Table \ref{tab:eventpage} describes the relevant content to be extracted from each event data source. In this regard, we implemented a comprehensive filtering and cleaning pipeline to isolate the target information from the raw data collection. 

Subsequently, the filtered event data from multiple pages were aggregated and structured into a standard JSON format featuring consistent fields across all seismic events, thereby ensuring uniformity in data representation. The fields in the JSON file are organized into textual, image and waveform categories. The text fields (e.g., "Regional information'',"Origin'') are in string format, the image fields (e.g., Intensity image, population exposure) contain file paths to PNG/JPEG images, and the waveform fields contain raw timeseries data in numerical format. A representative example of the resulting JSON structure for a particular event is shown in the Appendix under Listing 1.

\subsubsection{Spatial and Temporal coverage}

\begin{figure}[h!]
\centering
\includegraphics[width=0.48\textwidth]{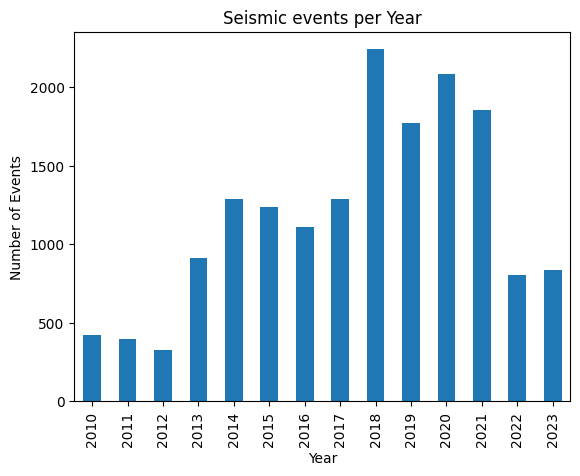}
\caption[short]{Frequency of recorded seismic events per year.}
\label{fig:diagram:temporal_distribution}
\end{figure}

The temporal distribution of the collected standardized data is shown in Figure \ref{fig:diagram:temporal_distribution}.
Fewer events occur in the early years of the data (2010-2012), steady from 2013-2017 but a considerable spike occurs between 2018-2021. This inference does not necessarily indicate an actual increase in global seismic activity but could be due to improved data reporting by the USGS (U.S. Geological Survey) and NEIC from the massive expansion of global seismic sensor networks and the inclusion of thousands of low-magnitude events (M2.0–M4.0).  

We visualize the spatial distribution of the data in Figure \ref{fig:diagram:spatial_distribution}.
Most of the events occur on the west side and east side of the Pacific Plate (Ring of Fire), as well as along the Australian Plate in the Indian Ocean and along the north sides of the Arabian and Indian Plates across south Asia.


\subsubsection{Text data}
The majority of the fields in the event JSON data contain textual content including ``Regional information'', ``Origin,'' ``Moment Tensor'', etc. We provide the mean length and the corresponding top words per field for some of the text fields to characterize the field's values in Table~\ref{tab:text_data_summary}.
The top words per event generally correspond to expected concepts, e.g. the top words for ``Event region'' reference geographic areas commonly represented in the data (``alaska'', ``california'').


\begin{table*}[t!]
\caption{Summary of the text data fields.}
\centering
\small
\scalebox{0.90}{
\begin{tabular}{ | p{2cm} | p{4cm} | r | p{6cm} | }  
\hline
\textbf{Field} & \textbf{Description} & \textbf{Mean (S.D.) tokens} & \textbf{Top-10 words} \\ \hline
Tectonic summary & Summary of historical tectonic activity in the event region. & 4169 (3115) & plate trench earthquakes new subduction australia along pacific earthquake arc \\ \hline
Event summary & Summary of event statistics and activity. & 1236 (212) & population earthquake estimated shaking exposure losses pager structures iii earthquakes \\ \hline
Event region & Brief description of event region. & 28 (10) & region islands alaska southern california coast northern new south near \\ \hline
\end{tabular}
}
\label{tab:text_data_summary}
\end{table*}

\subsubsection{Semi-structured data}

In addition to raw text, the dataset encompasses a wealth of key-value form structured data that can be leveraged for basic information extraction tasks, e.g. extracting the value of a particular field.
Shown in Table \ref{tab:json_data_summary}, each event's JSON file contain a large number of textual fields for a model to process, although many fields tend to be repeated across lists and dictionaries within the JSON file.
Most of the fields are string values, although some of the string values may also wrap numeric data, e.g., the numeric earthquake magnitude value is wrapped as the string ``$4.5\text{mlr} \pm 0.2$``.

\begin{table}[t!]
    \caption{Summary of JSON data.}
    \centering
    \begin{tabular}{ | l | r | }
        \hline
        \textbf{Statistic} & \textbf{Value} \\ \hline
        Total Fields & 1130 \\ \hline
        Unique Fields & 27.3 \\ \hline
        Proportion of String Values & 83.7\% \\ \hline
        Proportion of Numeric Values & 16.3\% \\ \hline
    \end{tabular}
    \label{tab:json_data_summary}
\end{table}

\begin{table}[t!]
    \caption{Event type distribution.}
    \centering
    \begin{tabular}{ | l | r | }
        \hline
        \textbf{Event type} & \textbf{Proportion} \\ \hline
        Earthquake & 99.7\% \\ \hline
        Volcanic eruption & 0.125\% \\ \hline
        Explosion  & 0.104\% \\ \hline
        Other & 0.0312\% \\ \hline
    \end{tabular}
    \label{tab:event_types}
\end{table}

One notable field in the JSON file is the type of seismic event, which can range from earthquakes to man-made explosions.
Shown in Table \ref{tab:event_types}, the majority of the events represent earthquakes, and a very small proportion of events are either eruptions or explosions.
This is due in part to the sampling procedure where we discarded events with a magnitude below 2.5, which includes many smaller events such as localized explosions.

\subsubsection{Images}

The images in the \multiseismo{} dataset fall into one of several categories:

\begin{itemize}
    \item Macroseismic intensity maps: Map images expressing felt intensity, shaking, and potential damage.
    \item Instrumental ground motion maps: Maps showing actual/estimated ground-motion parameters from the instruments.
    \item Ground motion attenuation/comparison plots: Scientific plots comparing measured ground motion data with model predictions.
\end{itemize}

\begin{figure*}[h!]
\centering
\includegraphics[width=0.95\textwidth]{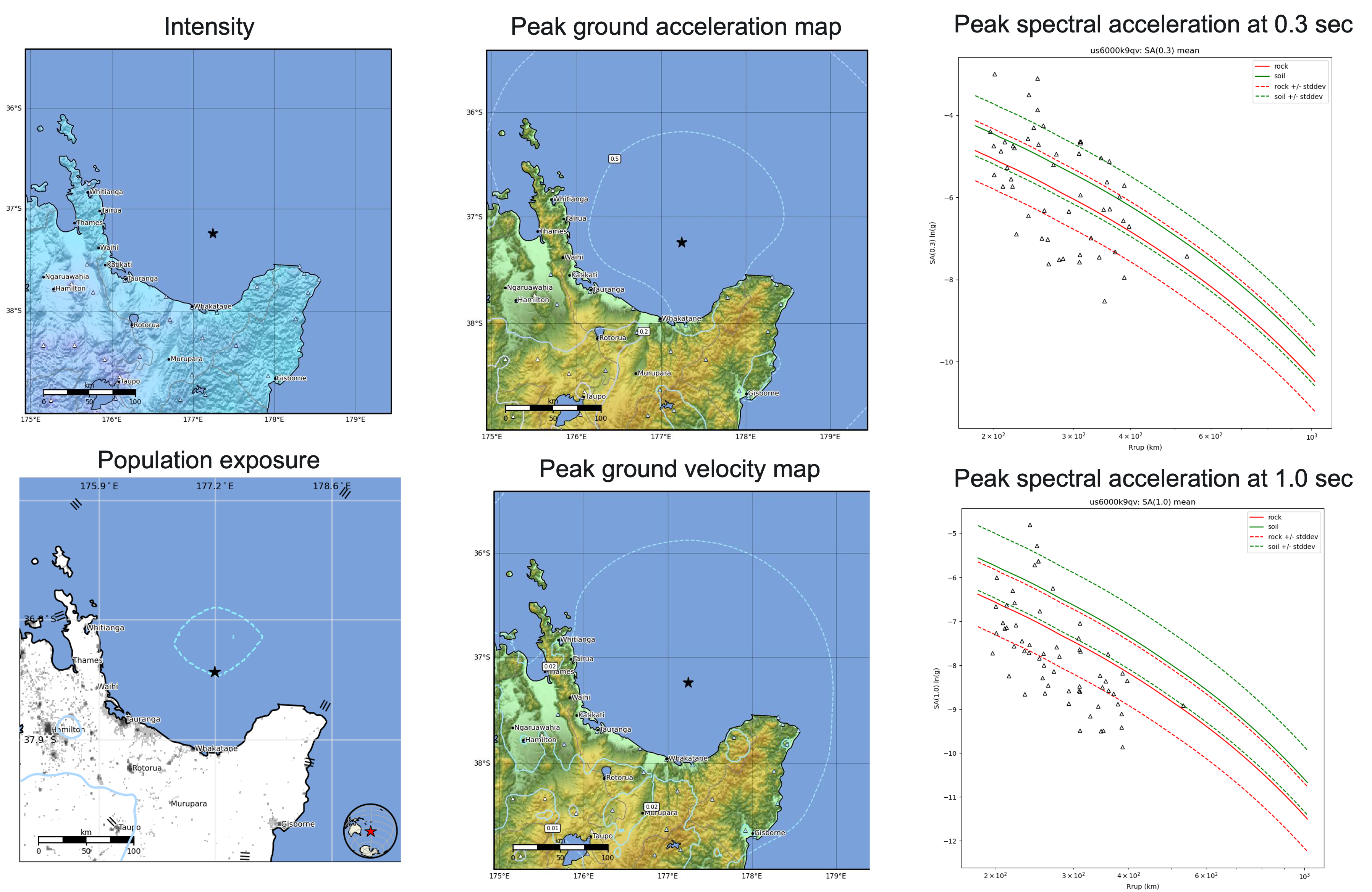}
\caption[short]{Example images, grouped by type. First column (Macroseismic intensity): Intensity and Population exposure. Second column (Instrumental ground motion maps): Peak ground acceleration and peak ground velocity. Third column (Ground motion attenuation/comparison plot): Peak spectral acceleration at 0.3 sec and Peak spectral acceleration at 1.0 sec.}
\label{fig:diagram:image_types}
\end{figure*}

\begin{table}[t!]
\caption{Image type distribution (N=91313 total).}
\centering
\begin{tabular}{ | p{5cm} | r | }
\hline
\textbf{Image type} & \textbf{Proportion} \\ \hline
Macroseismic intensity maps & 26.9\% \\ \hline 
Instrumental ground motion maps & 22.5\% \\ \hline 
Ground motion attenuation/comparison plots & 50.6\% \\ \hline 
\end{tabular}
\end{table}

Figure \ref{fig:diagram:image_types} denotes different image types. Most of the images are small- to medium-sized, ranging from 400-800 pixels wide and 200-900 pixels high.

\subsubsection{Time series}

\begin{table}[t!]
    \caption{Summary of time series data.}
    \centering
    \begin{tabular}{ | p{4.5cm} | r | }
        \hline
        \textbf{Statistic} & \textbf{Value (Mean)} \\ \hline
        Stations per event & 51.0 \\ \hline
        Stations near event (within 0.20 degrees) & 2.73 \\ \hline
        Channels per station & 2.80 \\ \hline
        Station distance from origin (degrees) & 4.42 \\ \hline
        Time length (sec) & 990 \\ \hline
        P-wave arrival time (sec) & 80.0 \\ \hline
        Zero-crossing rate & 0.0647 \\ \hline
        Peak period (sec) & 2.21 \\ \hline
    \end{tabular}
    \label{tab:time_data_summary}
\end{table}

To characterize the waveform data which is stored in time series format, we provide various statistics about the stations (at which the data were collected) and about the raw data in Table \ref{tab:time_data_summary}.
Notably, most events have a large number of stations recording the event, most stations have the expected number of channels (on average $3$ per station) and a relatively late p-arrival time.



\subsection{Instruction generation process}
One can directly leverage the \multiseismo{} dataset in its current form to train and evaluate GMMs in an unsupervised manner. However, to further show the utility of \multiseismo{}, we generated \micse{} (\textbf{M}ultimodal \textbf{I}nstructions to \textbf{C}haracterize \textbf{S}eismic \textbf{E}vents) based on common subject-area tasks that relate to specific modalities. 

\micse{} offers several additional benefits on top of the raw data: (i) it transforms unstructured seismic data into supervised learning tasks with explicit input-output pairs and training objectives; (ii) it establishes standardized benchmarking protocols for consistent model evaluation; and (iii) it facilitates cross-modal learning by requiring models to bridge information across modalities, connecting waveform patterns to textual metadata or integrating visual intensity maps with numerical data, thereby enabling comprehensive multimodal scientific reasoning.

In this regard, we worked with several seismology subject matter experts to identify modality-specific tasks that address common seismic knowledge collection needs and whose ground-truth answers can reasonably be extracted from the collected data with minimal extra effort and additional knowledge.
We developed a variety of question-focused instructions for each modality that vary in difficulty level and reasoning capability, ranging from information retrieval to mathematical reasoning.

We summarize the generated instructions in Table \ref{tab:instructions}. \micse{} comprises 17 instruction templates spanning text, image,and time series modalities. For each template, we generated instructions across all 16k events, yielding more than 500k total instructions.
\micse{} contains more text-modality instructions because it is easier to generate text-only instructions from the raw data, as compared to other modalities where the ground-truth answers are harder to extract (e.g., determining the number of stations shown on a map of the event).
We show example instructions and context for each modality in the Appendix (see Table~\ref{tab:example_instructions}).

\begin{table*}[t!]
\caption{Summary of example instructions generated for model training, per-modality.}
\centering 
\scalebox{0.92}{
\begin{tabular}{ | l | l | r | r | }
\hline
\textbf{Modality}    & \textbf{Instruction example}                                           & \# \textbf{Templates} & \textbf{\# Instructions} \\ \hline
Text        & What is the population of the nearest place to the epicenter? & 11           & 112,000         \\ \hline
Image       & Is the event epicenter on land or in water?                   & 3            & 14,000          \\ \hline
Time series & Is the event noise or a seismic event?                          & 3            & 67,000         \\ \hline
\end{tabular}
}
\label{tab:instructions}
\end{table*}



\section{Training and Evaluation}
\label{sec:evaluation}

The multimodal dataset developed in this work is intended to evaluate the seismic reasoning capabilities of multimodal language models.
We first evaluate several GMMs on the test set of the instruction dataset (MICSE) (\autoref{sec:instruction_evaluation}), which shows a surprising level of difficulty across the different modalities.
Next, we test the ability of a long-context LLM to respond to increasingly fine-grained time series (\autoref{sec:time_series_length_evaluation}), and we find that increasing the amount of time series data doesn't significantly change overall performance.
Lastly, we test the ability of multimodal encoder models to retrieve image and text data from the same event (\autoref{sec:multimodal_retrieval}) and show that the model with a more diverse training regime (ImageBind) outperforms the more limited-domain model (CLIP), although cross-modality retrieval remains difficult overall.

\subsection{Models used in instruction evaluation}

We evaluate the relative difficulty of MISCE by testing the performance of existing open-source multimodal LLMs.
The following modern multimodal models are evaluated, considering their wide adoption in the AI community:

\begin{itemize}
    \itemsep0em
    \item Unified-IO 2 ~\cite{lu2024unified},
    \item Phi 4 ~\cite{abouelenin2025phi4}.
\end{itemize}

Due to its modular design, the Unified-IO 2 model is relatively easy to modify, and we incorporate a time series encoder into its architecture to convert the raw time series data to a latent representation.
We add a pretrained Chronos-T5 encoder to the base Unified-IO 2 model~\cite{ansari2024chronos} and an accompanying projection layer.
In this way, the encoded time series data are projected to the correct number of hidden dimensions and passed to the combined LLM layer to be processed jointly with the encoded text data.

Furthermore, we train the modified Unified-IO 2 model on a held-out set of training data instructions to produce a model fine-tuned on the seismic domain, to assess the possible upper bound of model performance on this dataset.
The model is trained on text-only instructions, image instructions, and time series instructions in sequence.
We unfreeze only the modality-specific projection layers and the decoder LLM component of Unified-IO 2 under the assumption that those components are the most important in answering modality-specific questions.\footnote{The trained model weights will be made available on publication.}
The final model is called \underline{SeisModal}, to represent \underline{seis}mic multi\underline{modal} understanding.

Since the Phi-4 model cannot be easily modified to accommodate time series data, we provide the raw time series values as text to the Phi-4 model.
We rely on the model's long context window (128K tokens) to process the long seismic time series data effectively. 
Converting time series to text maintains the exact sequential order and numerical precision of sensor readings, which are critical for seismic analysis.
A single tokenized seismic waveform can exceed 400K tokens for 100K time steps, which requires us to down-sample the full time series to 1024 time steps for computational efficiency.

\subsection{Instruction evaluation}
\label{sec:instruction_evaluation}


\begin{table*}[t!]
    \caption{Model evaluation results on instructions from test data split, separated by modality. Bold values represent the best performance.}
    \centering
    \small
    \begin{tabular}{ | l | r r r | r r r | r r r | }
        \hline
        ~ & \multicolumn{3}{| c |}{\textbf{Text}} & \multicolumn{3}{| c |}{\textbf{Image}} & \multicolumn{3}{| c |}{\textbf{Time series}} \\ \hline
        \textbf{Model} & BLEU & ROUGE & BERT & BLEU & ROUGE & BERT & BLEU & ROUGE & BERT \\ \hline
        Phi 4 & 0.0299 & 0.0549 & \textbf{0.403} & 0.0924 & 0.124 & 0.491 & 0.0218 & 0.0407 & 0.389 \\ \hline
        Unified-IO & 0.0111 & 0.0243 & 0.350 & 0.545 & 0.501 & \textbf{0.786} & 0.0445 & 0.0187 & 0.352 \\ \hline
        \textbf{SeisModal} & \textbf{0.213} & \textbf{0.209} & 0.232 & \textbf{0.642} & \textbf{0.630} & 0.683 & \textbf{0.675} & \textbf{0.664} & \textbf{0.824} \\ \hline
    \end{tabular}    \label{tab:model_evaluation_results}
\end{table*}

We show the results for model performance on the test set of the multimodal instructions in \autoref{tab:model_evaluation_results}.
We measure model performance with standard text overlap metrics, including BLEU, ROUGE, and BERT score for semantic matching.

The following trends are identified in the overall results:
\begin{itemize}
    \item The \micse{} test data proves surprisingly difficult for both of the pretrained models, shown by the overall low scores across all modalities. This suggests an overall lack of seismic capability in current multimodal LLMs, which is not surprising considering the relative sparsity of seismic knowledge in general among likely training data (e.g., websites, common instruction-tuning datasets).
    \item The time series domain proves the most difficult for the pretrained models to understand, with especially low scores for the Phi model. However, the fine-tuned SeisModal model achieves much better results, particularly for the simple tasks such as distinguishing true seismic data from noise data. This suggests that an even more advanced model could be adapted to the time series domain with a similar procedure as the one applied to SeisModal, i.e. add a time-specific encoder to a general multimodal model and a corresponding projection layer.
    \item All models perform best in the image domain, possibly due to the more limited variation in image data as compared to other modalities (only geographic maps are included in \micse{}). The pretrained Unified-IO model outperforms the other baseline, possibly because the Phi-4 outputs highly verbose responses to the instructions which in turn lowers the text-overlap scores. Fine-tuning SeisModal in the image domain produces smaller gains than expected, and actually results in a decrease in the BERT score (relative to the baseline Unified-IO 2 model). The decrease may be due to the fine-tuned model learning to guess numbers more randomly on some of the numeric tasks (e.g. counting stations in an image), which may be fixed by training on a larger or less diverse dataset.
\end{itemize}

\subsection{Increasing time series length}
\label{sec:time_series_length_evaluation}

Seismic waveforms often contain complex patterns that indicate important real-world events such as p-wave arrival, which can only be identified through fine-grained analysis of the data.
To test the sensitivity of the long-context models, we provide increasingly fine-grained samples of the time series by increasing the sample rate from 1024 tokens up to 8092 tokens.
Intuitively, increasing the level of detail available to the models should provide a stronger signal for time series reasoning.
For simplicity, we evaluate only the time series task related to classifying a waveform as seismic or noise, and we evaluate the Phi-4 model due to its ability to handle long context input.

\begin{table}[t!]
    \caption{Evaluation on time series tasks with increasing time series sample size, using BLEU scores. N=660 data total over 100 sampled events.}
    \centering
    \begin{tabular}{ | l | *{4}{r|} }
        \hline
        ~ & \multicolumn{4}{c|}{\textbf{Time series length}} \\ \hline
        \textbf{Model} & \textbf{1024} & \textbf{2048} & \textbf{4096} & \textbf{8092} \\ \hline
        Phi 4 & 0.0159 & 0.0175 & 0.0190 & 0.0201 \\ \hline
    \end{tabular}
\label{tab:time_series_length_evaluation}
\end{table}

The results of the analysis are shown in Table \ref{tab:time_series_length_evaluation}.
Increasing the length of time series available slightly improves performance but with diminishing returns, such that the performance with the longest time series (8092) is only slightly better than the next-longest (4096).
The model's output also indicates significant uncertainty with the task, often including concerns such as ``the time series does not provide any context or metadata that would allow for a definitive classification.''
Rather than providing a stronger data signal for the model to use in reasoning, increasing the level of granularity in the time series did not conclusively improve the model's performance or alleviate its uncertainty about the task.

\subsection{Multimodal data retrieval}
\label{sec:multimodal_retrieval}


\begin{table}[t!]
    \caption{Multimodal event data retrieval results (N$=7887$).}
    \centering
    \small
    \scalebox{0.8}{
    \begin{tabular}{| l | r | r | r | r |}
        \hline
        ~ & \multicolumn{2}{c|}{\textbf{Image-image}} & \multicolumn{2}{c|}{\textbf{Text-image}} \\ \hline
        \textbf{Model} & \textbf{MRR} & \textbf{R@10} & \textbf{MRR} & \textbf{R@10} \\ \hline
        CLIP & 0.00840 & 0.0134 & 0.000889 & 0.000254 \\ \hline
        ImageBind & 0.0173 & 0.0336 & 0.00834 & 0.0198 \\ \hline
        Random & 0.000117 & 0.00 & 0.0000974 & 0.00 \\ \hline
    \end{tabular}
    }
\label{tab:retrieval_evaluation}
\end{table}

In addition to evaluating the models on MICSE, we use the raw data (\multiseismo{}) to test the ability of encoder-only models to \emph{retrieve} multimodal data.
We test the following models due to their popularity and different training paradigms:

\begin{itemize}
    \item CLIP (trained on images and text captions) ~\cite{radford2021learning};
    \item ImageBind (trained on images, videos, audio, sensor data, and text) ~\cite{girdhar2023imagebind}.
\end{itemize}

Formally, for a set of modality-specific data $\mathcal{D}$, we test the ability of an encoder model $M$ to retrieve relevant data $\mathcal{D}_{e}$ from event $e$ based on a single datum $d_{e}$.
E.g., the model may have to retrieve a particular geographic map of the event using other maps from the same event (i.e., image-image).
For the image modality, we use geographic maps of the event, and for the text modality we use the title of the event that indicates magnitude and location of the event, e.g., ``M 5.1 - 55 km ENE of La Paz, Philippines.''

The retrieval results are shown in Table \ref{tab:retrieval_evaluation}, from which the following patterns emerge:

\begin{itemize}
    \item ImageBind is the superior retrieval model, outperforming CLIP in image-image retrieval and text-image retrieval.
    \item All trained models significantly outperform the random baseline but still show significant room for improvement, particularly in the text-image retrieval task. This suggests that the information available in the event title (magnitude and location) are only weakly tied to the event images, and this information may require a model with stronger reasoning capabilities or simply may require more text information to describe the event.
    \item The task is generally difficult, considering the large dataset size (N=7887 events with valid images for retrieval). Future use of this data may restrict the images or events to facilitate retrieval, or break retrieval into easier sub-tasks. 
\end{itemize}

\section{Conclusions and Future directions}
\label{sec:conclusion}

We develop \multiseismo{}, a comprehensive multimodal seismic dataset that addresses critical gap in data resources for training and evaluating multimodal language models in seismological applications. By systematically aggregating and structuring data from public USGS and NEIC databases, we have created a unified multimodal dataset that integrates seismic waveforms, visual artifacts, and textual descriptions in a format optimized for multimodal AI. \multiseismo{} covers a wide temporal and spatial range, and the generated seismic multimodal instructions in \micse{} include complicated domain-specific tasks that require active reasoning and not merely memorization.
\revone{
Furthermore, we adapted and fine-tuned the existing Unified-IO 2 model with specialized time-series processing capabilities on \micse{} data to produce SeisModal, a domain-specific generalist multimodal model that demonstrates the feasibility of adapting general-purpose architectures for specialized scientific applications.}

Evaluating multimodal models on the instructions reveals a surprising level of difficulty even for well-supported modalities (text, image). The time series modality proves especially hard for pretrained models to understand even when provided with more fine-grained time data.
Using the data for multimodal retrieval similarly shows the limitations of models in understanding the connection between image and text in the seismic domain, which suggests the need for further domain-specific training for better performance. \revone{The evaluation performed shows that \multiseismo{} represents a rigorous benchmark with respect to the complexity of seismic analysis tasks, and the data can be used to measure progress in multimodal AI.
In parallel with the data, SeisModal serves as a proof-of-concept for domain-adapted multimodal systems in earth sciences.}

Future applications of \multiseismo{} include training specialized GMMs with enhanced time series processing capabilities, such as CNN-based waveform encoders or adversarial autoencoders, and developing unified representations for multi-station seismic recordings across varying distances. Additionally, \multiseismo{} can be integrated with existing multimodal datasets for domain-specific training, serve as a foundation for anomaly detection applications, and provide a template for SMEs in other scientific domains to develop similar multimodal instruction datasets using domain-specific imagery and time series data.

\section{Limitations}
\label{sec:limitations}

The main limitation of this datasets is the relatively small range of instruction templates for the image and time series domains (in Table \ref{tab:instructions}), particularly as it relates to the construction of \micse{}.
We found it surprisingly difficult to derive multimodal instruction templates that were both domain-specific and could be answered using the JSON data for the event.
E.g., questions about the intensity distribution shown in event images could not be formulated due to a lack of fine-grained information about event intensity, i.e., no ground-truth data about intensity at the level of individual cities.
Similarly, for the time series domain, most events had minimal data about waveform activity during the event outside than p-wave arrivals, which could have included other relevant data such as s-wave arrivals or surface waves.
Models trained on this dataset may require additional modality-specific instruction data to achieve desired performance in the seismic domain, depending on the model's purpose.
Secondly, for data storage purposes, we omit all events with less than 2.5 magnitude and therefore remove most human activity (e.g., mining) and non-earthquake events (volcanic eruptions), most of which have a relatively small magnitude.
Future work should expand this dataset to include smaller seismic events that may prove more challenging for LLMs to understand based on more subtle patterns in the data, e.g. smaller spikes indicating p-wave arrival.
A third limitation is the lack of explanations for the ground-truth answers, which may be useful for training interpretable reasoning models~\cite{lu2022learn}.
\section{Ethical Consideration}

It is unlikely that the datasets studied in this work will cause ethical concerns if released to the public, considering that the seismic data doesn't contain any sensitive information about individual people or organizations.
The datasets are intended primarily for evaluation and training of multimodal LLMs, and are intended for scientific, non-harmful applications such as post-hoc seismic data analysis.
However, we did not manually review all data included in \multiseismo{}, and if any harmful or sensitive information is shown to exist then we will take steps to remove it from the data.


\bibliographystyle{IEEEtran}
\bibliography{custom}

\newpage

\appendices
\newpage
\section{Example instructions}
\label{sec:exampleinstructions}

We show example instructions for text, image, and time series modality in Table~\ref{tab:example_instructions}.
Even the relatively simpler text modality still features somewhat complex instructions such as the request to identify the population of the nearest location to the epicenter, which requires (1) finding the nearest location, (2) extracting its population as a number.

\begin{table*}[t!]
\centering
\small
\begin{tabular}{l | p{4cm} | p{4cm} | p{4cm}}
Modality & Text                                                                                                                                                                                                                                                                                                                                                                                                                                                                                                                                                                                                                             & Image                                                  & Time series                                    \\ \hline
Prompt   & What is the population of the nearest place to the epicenter of the seismic event "M 5.7 - 140 km E of Laikit, Laikit II (Dimembe), Indonesia”?                                                                                                                                                                                                                                                                                                                                                                                                                                                                                  & Did the event epicenter occur on land or in the water? & Is the time series noise data or seismic data? \\ \hline
Context  & Document: \{"Administrative\_Region": \{\}, "Nearby\_Places": {[}\{"name": "Bitung, North Sulawesi, Indonesia", "distance": "123.3 km (76.6 mi) W", "population": "225134"\}, \{"name": "Tondano, North Sulawesi, Indonesia", "distance": "149.1 km (92.6 mi) W", "population": "33317"\}, \{"name": "Ternate, North Maluku, Indonesia", "distance": "152.8 km (95 mi) ESE", "population": "205001"\}, \{"name": "Manado, North Sulawesi, Indonesia", "distance": "154 km (95.7 mi) W", "population": "451916"\}, \{"name": "Tomohon, North Sulawesi, Indonesia", "distance": "160.8 km (99.9 mi) W", "population": "27624"\}{]} & \begin{minipage}{.25\textwidth} \includegraphics[width=\linewidth]{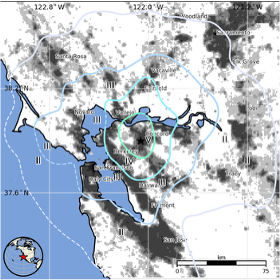} \end{minipage}                                           & \begin{minipage}{.25\textwidth} \includegraphics[width=\linewidth]{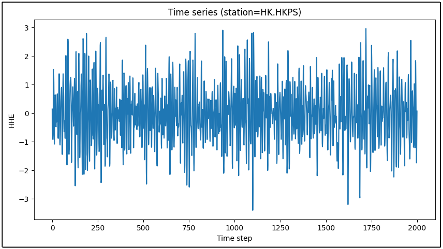} \end{minipage} (actual time series provided as raw numbers to model, not as visualization)                                    \\ \hline
Answer   & 225134                                                                                                                                                                                                                                                                                                                                                                                                                                                                                                                                                                                                                           & The epicenter occurred on land.                        & The time series data are seismic.             \\ \hline
\end{tabular}
\caption{Example instructions for text, image, and time series modalities.}
\label{tab:example_instructions}
\end{table*}

\newpage

\section{Sample data(JSON) file}
\label{sec:jsonfile}



\begin{lstlisting}[style=json, caption={JSON data structure.}, label=lst:json]
{
    "M 5.6 - 82 km NNE of Whakatane, New Zealand": {
        "metadata": {
            "id": "us6000k9qv",
            "title": "M 5.6 - 82 km NNE of Whakatane, New Zealand"
        },
        "Regional_Information": {
            "Administrative_Region": {},
            "Nearby_Places": [
                {
                    "name": "Whakatane, Bay of Plenty, New Zealand",
                    "distance": "82.8 km (51.5 mi) SSW",
                    "population": "18602"
                },
                {
                    "name": "Tauranga, Bay of Plenty, New Zealand",
                    "distance": "107.1 km (66.5 mi) WSW",
                    "population": "155200"
                },
                {
                    "name": "Rotorua, Bay of Plenty, New Zealand",
                    "distance": "132.9 km (82.6 mi) SW",
                    "population": "65901"
                },
                {
                    "name": "Tokoroa, Waikato, New Zealand",
                    "distance": "163.7 km (101.7 mi) SW",
                    "population": "14500"
                },
                {
                    "name": "Gisborne, Gisborne, New Zealand",
                    "distance": "170.5 km (105.9 mi) SSE",
                    "population": "37300"
                }
            ],
            "Tectonic_Summary": "\n Seismotectonics of the Eastern Margin of the Australia Plate"
        },
        "Origin": {
            "Phases": "us6000k9qv/origin/phase_data.csv",
            "Details": {
                "Magnitude": "5.6 mww\u00b1 0.0",
                "Location": "37.242\u00b0S177.243\u00b0E\u00b1 8.3 km",
                "Depth": "186.0 km\u00b1 1.9",
                "Origin Time": "2023-05-05 11:37:09.765 UTC",
                "Number of Stations": "61",
                "Number of Phases": "62",
                "Minimum Distance": "100.7 km(0.91\u00b0)",
                "Travel Time Residual": "0.86 s",
                "Azimuthal Gap": "49\u00b0",
                "FERegion": "off the east coast of the North Island of New Zealand (160)",
                "Review Status": "REVIEWED",
                "Catalog": "US( us6000k9qv )",
                "Location Source": "US1",
                "Magnitude Source": "US1",
                "Contributor": "US1"
            }
        },
        "Moment-tensor": {
            "W-phase_Moment_Tensor": {
                "Moment": "3.343e+17 N-m",
                "Magnitude": "5.62  Mww",
                "Depth": "180.5 km",
                "Percent DC": "75%",
                "Half Duration": "1.57 s",
                "Catalog": "US",
                "Data Source": "US 1",
                "Contributor": "US 1"
            },
            "Nodal_Planes": "[{'Plane': 'NP1', 'Strike': '248\u00b0',",
            "Principal_Axes": "[{'Axis': 'T', 'Value': '3.540e+17', 'Plunge': '66\u00b0', 'Azimuth': '260\u00b0'}"
        },
        "onepager": {
            "page-1": {
                "image-1": {
                    "image_ocr": "~G USGS : . science for a changing worl...",
                    "image_path": "us6000k9qv_onepager/figure-1-1.jpg"
                },
                "image-2": {
                    "image_ocr": "Earthquake Shaking Green Alert",
                    "image_path": "us6000k9qv_onepager/figure-1-2.jpg"
                },
                "image-3": {
                    "image_ocr": "",
                    "image_path": "us6000k9qv_onepager/figure-1-3.jpg"
                },
                "image-4": {
                    "image_ocr": "0 5 5 ie a Te SEN NY co) uy {Whitianga h ; 3 \\ phames « \" AL Waihi a * ...",
                    "image_path": "us6000k9qv_onepager/figure-1-4.jpg"
                },
                "table-1": "<table><tr><td>Fatalities 1,000</td><td>100 1,000 USD...</td><td>10</td></tr></table>",
                "table-2": "<table><thead><th colspan=\"2\" rowspan=\"2\">EXPOSURE (k=x1000) ESTIMATED MODIFIED MERCALLI...</th></thead></table>",
                "table-3": "<table><thead><th>Date (UTC)</th><th>Dist .. (truncated for brevity)",
                "table-4": "<table><thead><th>MMI</th><th>City</th><th> .. (truncated for brevity)",
                "text": "M 5.6, off the east coast of the North Island of New Zealand Origin Time:... (truncated for brevity)\n"
            }
        }
        "shakemap": {
            "intensity": "us6000k9qv/intensity.jpg",
            "pga": "us6000k9qv/pga.jpg",
            "pgv": "us6000k9qv/pgv.jpg",
            "psa0p3": "us6000k9qv/psa0p3.jpg",
            "psa1p0": "us6000k9qv/psa1p0.jpg",
            "psa3p0": "us6000k9qv/psa3p0.jpg",
            "psa0p3_regr": "us6000k9qv/psa0p3_regr.png",
            "psa1p0_regr": "us6000k9qv/psa1p0_regr.png",
            "psa3p0_regr": "us6000k9qv/psa3p0_regr.png",
            "info": {
                "multigmpe": {
                    "PGV": {
                        "gmpes": [
                            {
                                "gmpes": [
                                    "[AtkinsonBoore2003SSlab]",
                                    "[AtkinsonBoore2003SSlabCascadia]",
                                    "[ZhaoEtAl2016SSlab]",
                                    "[AbrahamsonEtAl2015SSlab]"
                                ],
                                "weights": [
                                    0.1667,
                                    0.1667,
                                    0.3333,
                                    0.3333
                                ],
                                "name": "subduction_slab_nshmp2014"
                            }
                        ],
                        "weights": [
                            1.0
                        ],
                        "name": "gmpe_us6000k9qv_custom"
                    },
                    "SA(0.3)": {
                        "gmpes": [
                            {
                                "gmpes": [
                                    "[AtkinsonBoore2003SSlab]",
                                    "[AtkinsonBoore2003SSlabCascadia]",
                                    "[ZhaoEtAl2016SSlab]",
                                    "[AbrahamsonEtAl2015SSlab]"
                                ],
                                "weights": [
                                    0.1667,
                                    0.1667,
                                    0.3333,
                                    0.3333
                                ],
                                "name": "subduction_slab_nshmp2014"
                            }
                        ],
                        "weights": [
                            1.0
                        ],
                        "name": "gmpe_us6000k9qv_custom"
                    },
                    "SA(1.0)": {
                        "gmpes": [
                            {
                                "gmpes": [
                                    "[AtkinsonBoore2003SSlab]",
                                    "[AtkinsonBoore2003SSlabCascadia]",
                                    "[ZhaoEtAl2016SSlab]",
                                    "[AbrahamsonEtAl2015SSlab]"
                                ],
                                "weights": [
                                    0.1667,
                                    0.1667,
                                    0.3333,
                                    0.3333
                                ],
                                "name": "subduction_slab_nshmp2014"
                            }
                        ],
                        "weights": [
                            1.0
                        ],
                        "name": "gmpe_us6000k9qv_custom"
                    },
                    "SA(3.0)": {
                        "gmpes": [
                            {
                                "gmpes": [
                                    "[AtkinsonBoore2003SSlab]",
                                    "[AtkinsonBoore2003SSlabCascadia]",
                                    "[ZhaoEtAl2016SSlab]",
                                    "[AbrahamsonEtAl2015SSlab]"
                                ],
                                "weights": [
                                    0.1667,
                                    0.1667,
                                    0.3333,
                                    0.3333
                                ],
                                "name": "subduction_slab_nshmp2014"
                            }
                        ],
                        "weights": [
                            1.0
                        ],
                        "name": "gmpe_us6000k9qv_custom"
                    },
                    "PGA": {
                        "gmpes": [
                            {
                                "gmpes": [
                                    "[AtkinsonBoore2003SSlab]",
                                    "[AtkinsonBoore2003SSlabCascadia]",
                                    "[ZhaoEtAl2016SSlab]",
                                    "[AbrahamsonEtAl2015SSlab]"
                                ],
                                "weights": [
                                    0.1667,
                                    0.1667,
                                    0.3333,
                                    0.3333
                                ],
                                "name": "subduction_slab_nshmp2014"
                            }
                        ],
                        "weights": [
                            1.0
                        ],
                        "name": "gmpe_us6000k9qv_custom"
                    }
                },
                "input": {
                    "event_information": {
                        "depth": "186.0",
                        "event_id": "us6000k9qv",
                        "eventsource": "us",
                        "netid": "us",
                        "eventsourcecode": "6000k9qv",
                        "id": "us6000k9qv",
                        "productcode": "us6000k9qv",
                        "productsource": "us",
                        "producttype": "shakemap",
                        "event_ref": "Origin",
                        "fault_ref": "Origin",
                        "intensity_observations": "0",
                        "latitude": "-37.242",
                        "longitude": "177.2433",
                        "location": "82 km NNE of Whakatane, Bay of Plenty, NZ",
                        "magnitude": "5.6",
                        "origin_time": "2023-05-05T11:37:09.000000Z",
                        "seismic_stations": "65",
                        "src_mech": "RS",
                        "event_description": "USGS NEIC ShakeMap",
                        "event_type": "ACTUAL"
                    }
                },
                "strec": {
                    "TectonicRegion": "Subduction",
                    "FocalMechanism": "RS",
                    "TensorType": "Mww",
                    "TensorSource": "us_6000k9qv",
                    "KaganAngle": 24.310755732291984,
                    "CompositeVariability": null,
                    "NComposite": 0,
                    "DistanceToStable": 257.7769676385941,
                    "DistanceToActive": 190.815526091277,
                    "DistanceToSubduction": 0,
                    "DistanceToVolcanic": null,
                    "Oceanic": false,
                    "DistanceToOceanic": 198.71477331026475,
                    "DistanceToContinental": 0,
                    "SlabModelRegion": "Kermadec-Tonga",
                    "SlabModelDepth": 96.81977081298828,
                    "SlabModelDepthUncertainty": 29.51973533630371,
                    "SlabModelDip": 45.791690826416016,
                    "SlabModelStrike": 219.61561584472656,
                    "SlabModelMaximumDepth": 47
                },
                "output": {
                    "ground_motions": {
                        "PGV": {
                            "units": "cms",
                            "bias": "0.0",
                            "max_grid": "0.504",
                            "max": "0.504"
                        },
                        "SA(0.3)": {
                            "units": "g",
                            "bias": "0.0",
                            "max_grid": "0.03",
                            "max": "0.028"
                        },
                        "SA(1.0)": {
                            "units": "g",
                            "bias": "0.0",
                            "max_grid": "0.006",
                            "max": "0.006"
                        },
                        "SA(3.0)": {
                            "units": "g",
                            "bias": "0.0",
                            "max_grid": "0.001",
                            "max": "0.001"
                        },
                        "PGA": {
                            "units": "g",
                            "bias": "0.0",
                            "max_grid": "0.009",
                            "max": "0.009"
                        },
                        "MMI": {
                            "units": "intensity",
                            "bias": "0.0",
                            "max_grid": "3.538",
                            "max": "3.456"
                        }
                    },
                    "map_information": {
                        "grid_points": {
                            "longitude": "543",
                            "latitude": "433",
                            "units": ""
                        },
                        "grid_spacing": {
                            "longitude": "0.0083333",
                            "latitude": "0.0083333",
                            "units": "degrees"
                        },
                        "grid_span": {
                            "longitude": "4.517",
                            "latitude": "3.6",
                            "units": "degrees"
                        },
                        "min": {
                            "longitude": "174.933",
                            "latitude": "-39.0",
                            "units": "degrees"
                        },
                        "max": {
                            "longitude": "179.45",
                            "latitude": "-35.4",
                            "units": "degrees"
                        }
                    },
                    "uncertainty": {
                        "grade": "--",
                        "mean_uncertainty_ratio": null,
                        "total_flagged_mi": "0",
                        "total_flagged_pgm": "22"
                    }
                },
                "processing": {
                    "ground_motion_modules": {
                        "gmpe": {
                            "module": "gmpe_us6000k9qv_custom",
                            "reference": ""
                        },
                        "ipe": {
                            "module": "VirtualIPE",
                            "reference": ""
                        },
                        "gmice": {
                            "module": "WGRW12",
                            "reference": ""
                        },
                        "ccf": {
                            "module": "LothBaker2013",
                            "reference": ""
                        },
                        "basin_correction": {
                            "module": "None",
                            "reference": ""
                        },
                        "directivity": {
                            "module": "None",
                            "reference": ""
                        }
                    },
                    "miscellaneous": {
                        "bias_max_dsigma": "1.5",
                        "bias_max_mag": "7.7",
                        "bias_max_range": "120.0",
                        "median_dist": "True",
                        "do_outliers": true,
                        "outlier_deviation_level": "3.0",
                        "outlier_max_mag": "7.0"
                    },
                    "shakemap_versions": {
                        "shakemap_revision": "4.0.2+731.g88a8033",
                        "shakemap_revision_id": "88a8033d4000490cbadd760afca0926926f510ed",
                        "process_time": "2023-05-14T05:29:54Z",
                        "map_version": 23,
                        "map_comment": "\"Origin updated\"",
                        "map_data_history": [
                            [
                                "2023-05-05T11:53:54Z",
                                "us",
                                1,
                                "\"Event added\""
                            ],
                            [
                                "2023-05-05T12:09:30Z",
                                "us",
                                2,
                                "\"Data association\""
                            ],
                        ],
                        "map_status": "automatic"
                    },
                    "site_response": {
                        "vs30default": "760.0",
                        "site_correction": "GMPE native"
                    }
                }
            },
            "rupture": {
                "type": "FeatureCollection",
                "metadata": {
                    "reference": "Origin",
                    "id": "us6000k9qv",
                    "network": "USGS National Earthquake Information Center, PDE",
                    "netid": "us",
                    "productcode": "us6000k9qv",
                    "time": "2023-05-05T11:37:09.000000Z",
                    "lat": -37.242,
                    "lon": 177.2433,
                    "depth": 186.0,
                    "mag": 5.6,
                    "locstring": "82 km NNE of Whakatane, Bay of Plenty, NZ",
                    "mech": "ALL",
                    "rake": 0.0
                },
                "features": [
                    {
                        "type": "Feature",
                        "properties": {
                            "rupture type": "rupture extent"
                        },
                        "geometry": {
                            "type": "Point",
                            "coordinates": [
                                177.2433,
                                -37.242,
                                186.0
                            ]
                        }
                    }
                ]
            }
        },
        "waveforms": {
            "7D.IVOR": {
                "metadata": {
                    "net_code": "[b'7D']",
                    "sta_code": "[b'IVOR']",
                    "sta_ele": "[997.]",
                    "sta_lat": "[-45.114782]",
                    "sta_lon": "[167.526721]"
                },
                "HH1": {
                    "data": "[-2584, -2590, -2566,]",
                    "metadata": {
                        "channel": "b'HH1'",
                        "delta": "0.01",
                        "end_time": "b'2023-05-05T11:49:38.930000Z'",
                        "npts": "66000",
                        "sampling_rate": "100.0",
                        "start_time": "b'2023-05-05T11:38:38.940000Z'"
                    }
                },
                "HH2": {
                    "data": "[-527, -516, -541,]",
                    "metadata": {
                        "channel": "b'HH2'",
                        "delta": "0.01",
                        "end_time": "b'2023-05-05T11:49:38.930000Z'",
                        "npts": "66000",
                        "sampling_rate": "100.0",
                        "start_time": "b'2023-05-05T11:38:38.940000Z'"
                    }
                },
                "HHZ": {
                    "data": "[434, 428, 430,]",
                    "metadata": {
                        "channel": "b'HHZ'",
                        "delta": "0.01",
                        "end_time": "b'2023-05-05T11:49:38.930000Z'",
                        "npts": "66000",
                        "sampling_rate": "100.0",
                        "start_time": "b'2023-05-05T11:38:38.940000Z'"
                    }
                }
            }
        }
    }
}
\end{lstlisting}





\end{document}